\documentclass[11pt,letterpaper]{article}

\usepackage[T1]{fontenc}
\usepackage[utf8]{inputenc}
\usepackage{lmodern}

\usepackage[margin=1in]{geometry}

\usepackage{amsmath,amssymb,mathtools,bm,accents,mathrsfs}

\usepackage{graphicx}
\usepackage{array}
\usepackage{booktabs}
\usepackage{threeparttable}
\usepackage{longtable}
\usepackage{multirow}
\usepackage{adjustbox}
\usepackage{float}
\usepackage{multicol}

\usepackage{authblk}
\usepackage{orcidlink}

\usepackage{pgfplots}
\usepackage{pgfplotstable}
\pgfplotsset{compat=1.18}

\usepackage{xcolor}
\usepackage{hyphenat}
\usepackage[most]{tcolorbox}
\usepackage[ruled,vlined]{algorithm2e}
\usepackage{appendix}
\usepackage{enumitem}

\usepackage{xurl}

\usepackage{hyperref}
\hypersetup{hidelinks}
\newcommand{\Algorithm}{Algorithm}

\tcbset{
  myalgorithm/.style={
    enhanced,
    colback=gray!5,
    colframe=black!70,
    boxrule=0.8pt,
    arc=4pt,
    left=8pt,
    right=8pt,
    top=6pt,
    bottom=6pt,
    fonttitle=\bfseries,
    coltitle=black,
    colbacktitle=gray!5,
    title={#1}
  }
}

\newtcolorbox{mybox}[1][]{
  enhanced jigsaw,
  colframe=black,
  colback=white,
  boxrule=0.4pt,
  sharp corners,
  left=4pt, right=4pt, top=4pt, bottom=4pt,
  frame code={
    \path[draw=black,line width=0.4pt]
      ([xshift=0.3pt,yshift=0.3pt]frame.south west)
      rectangle
      ([xshift=-0.3pt,yshift=-0.3pt]frame.north east);
  },
  interior code={
    \path[fill=white]
      ([xshift=0.3pt,yshift=0.3pt]frame.south west)
      rectangle
      ([xshift=-0.3pt,yshift=-0.3pt]frame.north east);
  },
  #1
}

\tcbset{before upper={\centering}}

\newcounter{myalgorithm}
\renewcommand{\themyalgorithm}{\arabic{myalgorithm}}

\title{MIBoost: A gradient boosting algorithm for variable selection after multiple imputation}

\author[1]{Robert Kuchen\thanks{Corresponding author: Robert Kuchen, Institute of Medical Biostatistics, Epidemiology and Informatics (IMBEI), University Medical Center, Johannes Gutenberg University Mainz, Mainz, Germany. Email: \texttt{robert.kuchen@uni-mainz.de}}\,\orcidlink{0000-0003-4384-7511}}
\author[1]{Noemi Castelletti\,\orcidlink{0000-0002-6598-5352}}
\author[1]{Konstantin Strauch\,\orcidlink{0000-0002-1814-5860}}

\affil[1]{Institute of Medical Biostatistics, Epidemiology and Informatics (IMBEI), University Medical Center, Johannes Gutenberg University Mainz, Mainz, Germany}

\date{}

\begin{document}

\maketitle

\begin{abstract}
Statistical learning methods for automated variable selection, such as the Least Absolute Shrinkage and Selection Operator (LASSO), elastic nets, and gradient boosting, have become increasingly popular tools for building powerful prediction models. Yet, in practice, analyses are often complicated by missing data. The most widely used approach to address missingness is multiple imputation, which involves creating several completed datasets. However, there is an ongoing debate about how to perform model selection in the presence of multiple imputed datasets. Simple strategies, such as pooling models across datasets, have been shown to have suboptimal properties. Although more sophisticated methods exist, they are often difficult to implement and therefore not widely applied. In contrast, two recent approaches extend the regularization methods LASSO and elastic nets to multiply imputed datasets by defining a single loss function, resulting in a unified set of coefficients across imputations. Our key contribution is to extend this principle to the framework of component-wise gradient boosting by proposing \texttt{MIBoost}, a novel algorithm that employs a uniform variable-selection mechanism across imputed datasets, together with its corresponding cross-validation routine \texttt{MIBoostCV}. In a simulation study, \texttt{MIBoost} yielded predictive performance comparable to that of other established methods, providing a practical boosting-based approach for variable selection with multiply imputed data. The proposed framework is implemented as the \texttt{R} package \texttt{booami}.
\end{abstract}

\begin{center}
\begin{minipage}{0.78\textwidth}
\small
\noindent\textbf{Keywords:} Gradient boosting; Multiple imputation; Variable selection; Model selection; Component-wise gradient boosting; Prediction; Missing data.
\end{minipage}
\end{center}
\medskip

\noindent\begin{minipage}{\textwidth}
\medskip
\noindent\textbf{Abbreviations}
\par\smallskip
\noindent
\begin{tabular}{@{}>{\raggedright\arraybackslash}p{0.14\textwidth}>{\raggedright\arraybackslash}p{0.31\textwidth}@{\hspace{1.5em}}>{\raggedright\arraybackslash}p{0.14\textwidth}>{\raggedright\arraybackslash}p{0.31\textwidth}@{}}
CV & cross-validation & \texttt{MIBoost} & Multiple-Imputation Gradient Boosting \\
EA & estimate averaging & \texttt{MIBoostCV} & cross-validation framework for \texttt{MIBoost} \\
EM & expectation--maximization & \texttt{SaENET} & stacked adaptive elastic net \\
LASSO & Least Absolute Shrinkage and Selection Operator & \texttt{SaLASSO} & stacked adaptive LASSO \\
MAR & missing at random & \texttt{SENET} & stacked elastic net \\
MI & multiple imputation & TNP & true negative proportion \\
MICE & multivariate imputation by chained equations & TPP & true positive proportion \\
MSPE & mean squared prediction error & & \\
\end{tabular}
\par\medskip
\end{minipage}

\raggedbottom

\section{Introduction}
Multiple imputation is widely considered the gold standard for handling missing data. When the main interest lies in causal inference, coefficients and standard errors obtained from each imputed dataset can be easily aggregated using Rubin’s pooling rules \cite{Rubin.1987}. However, performing variable selection for prediction purposes with multiple imputed datasets remains methodologically challenging, as the selected models can differ substantially across datasets. Many previously proposed methods for obtaining a single unified model are complex and insufficiently implemented in software packages. Simply selecting all variables chosen in any of the datasets often leads to overly complex models \cite{Wood.2008}; hence, many users resort to ad hoc solutions, such as including only variables selected in a minimum proportion (e.g., 50\%) of imputed datasets. While this heuristic may yield satisfactory results in some cases, it suffers from a strong dependence on an arbitrary threshold and can be unstable in small samples or in the presence of correlated predictors \cite{Zhao.2017}. To address this problem, two frameworks—grouping and stacked methods—have been proposed in recent years, in which a single loss function is minimized, ultimately yielding a unified model. Building on prior work in \cite{Chen.2013,Wan.2015}, algorithms were developed that incorporate the popular variable-selection methods LASSO \cite{Tibshirani.1996} and the elastic net \cite{Zou.2005} into these frameworks, and they were implemented in the \texttt{R} package \texttt{miselect} \cite{Du.2022,Rix.2020}.

Another statistical-learning method often considered an alternative to LASSO and elastic nets is component-wise gradient boosting \cite{Buhlmann.2003}, which also induces coefficient shrinkage and variable selection and is therefore well suited for high-dimensional data. In contrast to standard LASSO and elastic-net approaches, which are typically formulated for generalized linear models with a prespecified design matrix, boosting can be combined with a wide range of loss functions and base-learners, allowing nonlinear effects to be incorporated more naturally. However, no method to date ensures uniform variable selection when applying component-wise gradient boosting to multiply imputed data. This paper extends the idea of defining a single loss function across imputations and proposes a novel algorithm for component-wise gradient boosting that enables consistent variable selection across all imputed datasets. This approach is implemented in the \texttt{R} package \texttt{booami} to enable accessibility for methodologically oriented as well as application-oriented researchers. \\

\noindent\textbf{Our main contributions are as follows:}
\begin{itemize}[leftmargin=1.5em]
\item We propose \texttt{MIBoost}, a novel boosting algorithm specifically designed for prediction after multiple imputation, which couples base-learner selection across imputed datasets to improve model stability and interpretability.
\item We develop a cross-validation strategy for \texttt{MIBoost} that avoids data leakage by separating training and validation imputations and aggregates performance across imputations in a principled way.
\item We conduct a comprehensive simulation study assessing the predictive accuracy and variable-selection performance of \texttt{MIBoost} compared to existing boosting-based and imputation-based approaches.
\item We provide the R package \texttt{booami} \cite{Kuchen.2025} that implements our proposed approach.
\end{itemize}

\vspace{6pt}

\section{Materials and methods}
\label{sec:mat_met}

\subsection{Background: Existing approaches and limitations}
\label{subsec:existing}
Let $\mathcal{D}=\{(\boldsymbol{x}_i,y_i)\}_{i=1}^n$ denote the dataset, in which $y_i$ is observed and $\boldsymbol{x}_i\in\mathbb{R}^p$ may contain missing values.
Multiple imputation (MI) yields $M$ completed datasets, indexed by $m=1,\dots,M$, with imputed covariates $\boldsymbol{x}_i^{(m)}$.

Two widely used post-imputation variable-selection strategies are \emph{estimate averaging} and \emph{selection-frequency thresholding}.
In estimate averaging, the model is fitted separately in each imputed dataset and the resulting coefficient estimates $\{\hat\beta^{(m)}\}_{m=1}^M$ are averaged as $\bar\beta = M^{-1}\sum_{m=1}^M \hat\beta^{(m)}$ \cite{Wood.2008}.
In selection-frequency thresholding, the model is fitted in each imputed dataset and only covariates selected in at least a prespecified proportion of imputed datasets (e.g., majority vote) are retained \cite{Zhao.2017}.
Both approaches can yield non-uniform selection across imputed datasets and may require ad hoc reconciliation to obtain a single final model. 

More recently, stacked approaches have been proposed in which a single penalized model is fitted on a weighted stacked dataset, with weights chosen so that each subject's $M$ copies sum to one \cite{Wan.2015}.
Within this framework, stacked adaptive LASSO (\texttt{SaLASSO}) and stacked adaptive elastic net (\texttt{SaENET}), proposed by \cite{Du.2022}, are also used here as simulation comparators.
For a broader overview of model-selection approaches with multiply imputed data, see Appendix~\ref{subsec:existing_approaches}.

\subsection{Component-wise gradient boosting: Notation}
\label{subsec:cw_gradient_boosting}
In component-wise gradient boosting, an additive predictor
$\eta(\boldsymbol{x})=\sum_{r=1}^p h_r(x_r)$ is fitted
by approximately minimizing the empirical risk
$\hat R(\eta)=\sum_{i=1}^n \rho\!\big(y_i,\eta(\boldsymbol{x}_i)\big)$
for a differentiable loss $\rho$ \cite{JeromeH.Friedman.2001,Buhlmann.2003}.
At iteration $t$, pseudo-residuals (negative gradients) are
\[
u_i^{[t]}=
-\left.\frac{\partial}{\partial \eta}
\rho\!\big(y_i,\eta\big)\right|_{\eta=\hat\eta^{[t-1]}(\boldsymbol{x}_i)}.
\]
For each base-learner $r\in\{1,\dots,p\}$ with function space $\mathcal{H}_r$, we fit
\[
\hat h_{\beta_r}^{[t]}=\arg\min_{h\in\mathcal{H}_r}\sum_{i=1}^n\big(u_i^{[t]}-h(x_{i,r})\big)^2,
\]
select
\[
r^*=\arg\min_{r=1,\dots,p}\sum_{i=1}^n\big(u_i^{[t]}-\hat h_{\beta_r}^{[t]}(x_{i,r})\big)^2,
\]
and update
\[
\hat\eta^{[t]}(\boldsymbol{x})=\hat\eta^{[t-1]}(\boldsymbol{x})+\nu\,\hat h_{\beta_{r^*}}^{[t]}(x_{r^*}),
\]
with step length $\nu\in(0,1]$ and early stopping at $t_{\mathrm{stop}}$, chosen by cross-validation (CV).

In the multiple-imputation setting, boosting quantities carry the superscript $(m)$, e.g., pseudo-residuals $u_i^{(m)[t]}$ and fitted base-learner functions $\hat h_{\beta_r}^{(m)[t]}$.
We define the within-imputation residual sum of squares
\[
L_r^{(m)[t]}:=\sum_{i=1}^n\Big(u_i^{(m)[t]}-\hat h_{\beta_r}^{(m)[t]}(x_{i,r}^{(m)})\Big)^2,
\]
and the pooled quantity across imputations,
\[
L_r^{[t]}:=\sum_{m=1}^M L_r^{(m)[t]}.
\]

In \texttt{MIBoost}, base-learner selection is coupled across imputed datasets during boosting to directly yield a single final model. For a more detailed introduction to component-wise gradient boosting, see Appendix~\ref{subsec:background_gradient_boosting}.

\subsection{Proposed method: \texttt{MIBoost}}
\label{subsec:mi_boost}

The core idea of \texttt{MIBoost} is that although separate gradient boosting algorithms are run on each of the $M$ imputed datasets, base-learner selection at each iteration is based on the \textbf{aggregated loss}, i.e., the sum of the losses across all datasets. This enforces a uniform selection mechanism for base-learners, while gradient computation and coefficient updates remain dataset-specific. The final model is obtained by averaging the fitted additive predictors across the imputed datasets, yielding a single combined predictor.

\vspace{8pt}

\noindent\Algorithm~\ref{alg:MIBoost} provides a mathematical formulation of our procedure. 
We assume that \(M\) imputed datasets are available, each consisting of covariates that have been centered component-wise using, for each covariate, a common mean across all imputations rather than imputation-specific means (see Algorithm~\ref{alg:MIBoostCV} for the exact centering scheme), and a common outcome vector \(\boldsymbol{y}\).
Superscripts $(m)$ and $[t]$ denote the imputation index ($m = 1,\dots,M$) and the boosting iteration ($t = 1,\dots,t_{\mathrm{stop}}$), respectively, where $(m)$ indicates that the corresponding quantity is computed on the $m$-th imputed dataset. 
Observations with a missing outcome $y_i$ are assumed to have been excluded prior to imputation, so that \(\boldsymbol{y}\) does not depend on $m$. Each imputed dataset $m$ maintains its own additive predictor $\hat{\eta}^{(m)[t]}$, initialized with either the mean of \(\boldsymbol{y}\) or zero. 
The procedure consists of four steps:

\vspace{8pt}

\hrule

\vspace{6pt}

\begin{enumerate}
    \renewcommand{\labelenumi}{\textbf{(\theenumi)}} %

    \item At the start of each boosting iteration, we compute the negative gradient vector (pseudo-residuals) $\boldsymbol{u}^{(m)}$ for each imputed dataset $m$. This vector is obtained as the negative derivative of the loss function with respect to the additive predictor, evaluated at the previous iteration’s predictor $\hat{\eta}^{(m)[t-1]}$, indicating the direction of the steepest descent in the loss.

    \item These pseudo-residuals $\boldsymbol{u}^{(m)}$ are then regressed on each base-learner $h_{\beta_r}$ (\(r \in \{1, \dots, p\}\)) independently within each imputed dataset, yielding dataset-specific fitted base-learner functions and corresponding residual sums of squares.  

    \item For each base-learner, the residual sums of squares are summed across the imputed datasets to obtain a single loss value $L_r^{[t]}$. The base-learner $r^*$ with the lowest aggregated loss in this iteration is then selected. This joint selection step ensures that the same model component is updated in all imputed datasets, improving stability and favoring predictors that are consistently informative across datasets.

    \item Finally, the additive predictors $\hat{\eta}^{(m)[t]}$ in each imputed dataset are updated by adding the dataset-specific fitted contribution of the selected base-learner $r^*$, scaled by the step length $\nu$ (typically $\nu = 0.1$), to the additive predictor from the previous iteration.
\end{enumerate}

\vspace{12pt}
\hrule
\vspace{12pt}

\addtocounter{enumi}{1} 
\noindent\textbf{(\theenumi) Post-processing:} The above procedure is repeated for \(t_{\mathrm{stop}}\) iterations.
After boosting has concluded, the additive predictors are averaged across imputed datasets to produce a single coherent final predictor. 

\[
\bar{\eta}(\boldsymbol{x})
:= \frac{1}{M} \sum_{m=1}^M \hat{\eta}^{(m)[t_{\mathrm{stop}}]}(\boldsymbol{x})
=: \hat{\eta}(\boldsymbol{x}).
\]
Although it is also possible not to pool the models and instead retain \(M\) different predictors, the primary interest usually lies in obtaining a single model. Accordingly, we consider only the pooled model.
 \texttt{MIBoost} is implemented in the function
\texttt{impu\_boost} of the \texttt{R} package \texttt{booami} \cite{Kuchen.2025}, which also provides
implementations of estimate averaging and selection-frequency thresholding---as defined in
Subsection~\ref{subsec:existing_approaches}---for gradient boosting, enabling direct comparison.

\vspace{24pt}

\subsection{Cross-validation framework for \texttt{MIBoost}: \texttt{MIBoostCV}}
\label{subsec:MIBoostCV}
To select the optimal stopping iteration \( t_{\mathrm{stop}}^* \) and evaluate out-of-sample performance, \(K\)-fold CV can be employed. A detailed mathematical formulation of this procedure is provided in Algorithm~\ref{alg:MIBoostCV}. The steps of the CV procedure are as follows:

\vspace{12pt}

\textbf{Outer loop over CV folds:}

\vspace{12pt}

\hrule

\vspace{6pt}

\begin{enumerate}
    \renewcommand{\labelenumi}{(\theenumi)} 
    \item   \textbf{split data into training and validation subset:}
    Omit all observations with missing values in the outcome variable \(\boldsymbol{y}\). The remaining dataset with missing values in the design matrix $\boldsymbol{X}$ is first partitioned into \(K\) disjoint folds. In each CV iteration, one fold serves as the validation subset, and the remaining \(K-1\) folds form the training subset.
    
\item \textbf{Multiple imputation:}
In each CV iteration \(k\), the training subset is multiply imputed \(M\) times using the chosen imputation procedure,
which imputes missing values in \(\boldsymbol{X}\) using only \(\boldsymbol{X}\) (never \(\boldsymbol{y}\)), producing \(M\)
complete training subsets. Crucially, to avoid data leakage, the imputation is performed \emph{after} the training--validation
split. The associated validation subset is then imputed \(M\) times using the \(M\) imputation models that had been fitted on
the training subset, ensuring consistency between training and validation imputations. For every imputed training subset, there
is thus a corresponding imputed validation subset.
    
\item \textbf{Centering:}
Subsequently, since boosting performs best with centered inputs, covariates are centered using grand training means (averaged across the \(M\) imputed training datasets). The same grand means are then used to center covariates in the \(M\) imputed validation datasets.
{\tiny }

\item \textbf{Inner loop over boosting iterations:}  
\begin{enumerate}[label=(4.\arabic*), leftmargin=2.5em]
    \item \textbf{Application of \texttt{MIBoost}:} 
    Within each fold, the \texttt{MIBoost} algorithm is applied to the \(M\) imputed and centered training subsets for \(t_{\mathrm{stop}}\) iterations, with base-learner selection coupled across imputed datasets as described in Algorithm \ref{alg:MIBoost}.
\item \textbf{Average additive predictors across imputations:}
At the end of each boosting iteration \(t \le t_{\mathrm{stop}}\), the imputation-specific additive predictors
\(\hat{\eta}^{(-k,m)[t]}\) are averaged across \(m=1,\dots,M\) to obtain a fold-specific pooled predictor
\(\overline{\eta}^{(-k)[t]}\).

\item \textbf{Prediction and error computation:}
The pooled predictor \(\overline{\eta}^{(-k)[t]}\) is evaluated on each of the \(M\) imputed validation subsets, and the
resulting validation errors are averaged across \(m=1,\dots,M\), yielding one fold-\(k\) error per iteration \(t\).
This fold-\(k\) error is stored for \(t=1,\dots,t_{\mathrm{stop}}\).
\end{enumerate}

\setcounter{enumi}{4} 
\item \textbf{Average prediction errors across folds:} Since the CV procedure involves \(K\) iterations, we obtain \(K\) averaged prediction errors for each boosting iteration. These are then averaged across CV folds to obtain a single final averaged prediction error for each iteration $t \leq  t_{\mathrm{stop}}$.

 \vspace{8pt}

\hrule

\vspace{8pt}

\item \textbf{Select optimal stopping iteration:}  Select \( t_{\mathrm{stop}}^* \) as the boosting iteration that is associated with the smallest final prediction error.

\item \textbf{Final model fitting:}
Impute the full dataset with missing values \(M\) times using the chosen imputation procedure, which imputes missing values in
\(\boldsymbol{X}\) using only \(\boldsymbol{X}\) (never \(\boldsymbol{y}\)), resulting in \(M\) complete training datasets. Subsequently, compute grand covariate means by averaging across the \(M\) imputed datasets and center each dataset using these means. Then, run \texttt{MIBoost} on these datasets for
\(t_{\mathrm{stop}}^*\) boosting iterations to obtain the final model.

\end{enumerate}

The CV procedure is implemented in the \texttt{R} package \texttt{booami} \cite{Kuchen.2025}. Specifically, it is available through the function \texttt{cv\_boost\_imputed} when the user provides data that have already been split and imputed, and through the function \texttt{cv\_boost\_raw} when starting from a single dataset with missing values. Although described here for \texttt{MIBoost}, this CV framework is generally applicable to cross-validation with missing data. In \texttt{booami}, it is implemented not only for \texttt{MIBoost} but also for estimate averaging and selection-frequency thresholding.

\refstepcounter{myalgorithm}\label{alg:MIBoost}
\begin{tcolorbox}[
  myalgorithm={Algorithm \themyalgorithm: \texttt{MIBoost} – \small{Multiple-Imputation Gradient Boosting}},
  % make the "definitions" part smaller ...
  fontupper=\small,
  boxsep=1mm, left=1mm, right=1mm, top=0.5mm, bottom=0.5mm
]
% --- local spacing/font tweaks so the larger remainder still fits on one page ---
\begingroup
\linespread{0.97}\selectfont

% display math spacing
\setlength{\abovedisplayskip}{2pt}
\setlength{\belowdisplayskip}{2pt}
\setlength{\abovedisplayshortskip}{2pt}
\setlength{\belowdisplayshortskip}{2pt}
\setlength{\jot}{2pt}

% multicols spacing
\setlength{\premulticols}{0pt}
\setlength{\postmulticols}{0pt}
\setlength{\multicolsep}{1pt}
\setlength{\columnsep}{10pt}

\textbf{The algorithm requires the following to be specified:} 
\begin{multicols}{2}
\begin{itemize}[leftmargin=*,nosep,topsep=0pt]
    \item $m \in \{1,\dots,M\}$: index of imputations.
    \item $i \in \{1,\dots,n\}$: index of observations.
    \item \(\{\mathcal{D}^{(m)}\}_{m=1}^M\): set of \(M\) imputed (and centered) datasets, where
    \(\mathcal{D}^{(m)}=\{(\boldsymbol{x}_i^{(m)},y_i)\}_{i=1}^n\).
    \item $p$: number of base-learner (candidate covariates).
    \item $t_{\mathrm{stop}}$: stopping iteration.
    \item $\nu$: step length (typically $\nu = 0.1$).
    \item $\rho$: differentiable loss function.
    \item $\mathcal{H}_r$: function space for base-learner $r$;
      \(\hat{h}_{\beta_r}^{(m)[t]} \in \mathcal{H}_r\) denotes the fitted base-learner \emph{update}
      for covariate \(r\) at iteration \(t\) in imputation \(m\).
    \item $\hat{\eta}^{(m)[t]}(\boldsymbol{x})$: additive predictor function for imputation $m$ at iteration $t$ (with \(\hat{\eta}^{(m)[0]}(\boldsymbol{x})\) as initialization).
\end{itemize}
\end{multicols}

\emph{All covariates are assumed centered (as in Algorithm \texttt{MIBoostCV}); for readability, we omit the bar and write \(\boldsymbol{x}\) for centered covariates throughout.}

% ... and make everything after the above sentence larger
\normalsize

\noindent\rule[0.5ex]{\linewidth}{0.5pt}
\textbf{Initialization:} For each \(m\), set \(\hat{\eta}^{(m)[0]}\) to the loss-minimizing constant (or an offset).

\noindent\rule[0.5ex]{\linewidth}{0.5pt}
\textbf{Iterative procedure (for $t = 1$ to $t_{\mathrm{stop}}$):}
\begin{enumerate}[leftmargin=*,itemsep=0pt,topsep=0pt,parsep=0pt]
    \item \textbf{Compute negative gradients:} For each imputation $m$ and observation $i$,
\[
u_i^{(m)[t]}
= - \left. \frac{\partial}{\partial \eta} \rho\!\big(y_i, \eta\big)
   \right|_{\eta = \hat{\eta}^{(m)[t-1]}(\boldsymbol{x}_i^{(m)})}.
\]

    \item \textbf{Fit base-learners:} For each $r \in \{1,\dots,p\}$ and $m$,
\[
\hat{h}_{\beta_r}^{(m)[t]}
= \arg\min_{h \in \mathcal{H}_r} \sum_{i=1}^n \big(u_i^{(m)[t]} - h(x_{i,r}^{(m)})\big)^2.
\]

    \item \textbf{Select best base-learner:} Compute pooled residual sums of squares and select $r^*$
\[
L_r^{[t]} := \sum_{m=1}^M \sum_{i=1}^n \Big(u_i^{(m)[t]} - \hat{h}_{\beta_r}^{(m)[t]}(x_{i,r}^{(m)})\Big)^2,
\qquad r^* = \arg\min_{r} L_r^{[t]}.
\]

    \item \textbf{Update predictors:} For each $m$, update the additive predictor function
\[
\hat{\eta}^{(m)[t]}(\boldsymbol{x}) := \hat{\eta}^{(m)[t-1]}(\boldsymbol{x}) 
  + \nu \, \hat{h}_{\beta_{r^*}}^{(m)[t]}(x_{r^*}).
\]
\end{enumerate}

\noindent\rule[0.5ex]{\linewidth}{0.5pt}
\begin{enumerate}[resume,leftmargin=*,itemsep=0pt,topsep=0pt,parsep=0pt]
    \item \textbf{Post-processing:} Average additive predictors across imputations:
\[
\bar{\eta}(\boldsymbol{x})
:= \frac{1}{M} \sum_{m=1}^M \hat{\eta}^{(m)[t_{\mathrm{stop}}]}(\boldsymbol{x})
=: \hat{\eta}(\boldsymbol{x}).
\]
\end{enumerate}

\endgroup
\end{tcolorbox}

\pagebreak

\refstepcounter{myalgorithm}\label{alg:MIBoostCV}
\begin{tcolorbox}[myalgorithm={Algorithm \themyalgorithm: \texttt{MIBoostCV} – \small{Cross-validation framework for \texttt{MIBoost}}}, breakable]

{\footnotesize
\textbf{The algorithm requires the following to be specified:}  
\begin{multicols}{2}
\begin{itemize}[leftmargin=*,nosep]
    \small
    \item $k \in \{1,\dots,K\}$: index of folds.
    \item $m \in \{1,\dots,M\}$: index of imputations.
    \item $\mathcal{D} = \{(\boldsymbol{x}_i, y_i)\}_{i=1, \ldots, n}$: full dataset with $\boldsymbol{x}_i \in \mathbb{R}^{p \times 1}$.
    \item $\mathcal{X}=\{\boldsymbol{x}_i\}_{i=1}^n$: covariate data (possibly missing).
    \item \parbox[t]{\linewidth}{$\mathcal{I}_k$: indices of observations in the validation subset of fold $k$.}
    \item $\mathrm{MIBoost}(\cdot)$: boosting procedure.
    \item $t_{\mathrm{stop}}$: maximum number of boosting iterations.
    \item $\mathrm{Imp}(\cdot)$: multiple-imputation procedure for covariates only,
    i.e.\ $\mathrm{Imp}:\mathcal{X}\mapsto\tilde{\mathcal{X}}$ (no outcome information is used).
\end{itemize}
\end{multicols}
}

\vspace{-10pt}

\noindent\rule[0.5ex]{\linewidth}{0.4pt}

\textbf{Outer loop: For each fold $k = 1,\dots,K$:}
\begin{enumerate}[leftmargin=*]
    \item \textbf{Split data into  training and validation subset:}
    \[
    \mathcal{D}^{(-k)} = \{(\boldsymbol{x}_i, y_i) : i \notin \mathcal{I}_k\}, \quad 
    \mathcal{D}^{(k)}  = \{(\boldsymbol{x}_i, y_i) : i \in \mathcal{I}_k\}.
    \]
    Let
    \[
      \mathcal{X}^{(-k)}=\{\boldsymbol{x}_i : i\notin\mathcal{I}_k\},\qquad
      \mathcal{X}^{(k)}=\{\boldsymbol{x}_i : i\in\mathcal{I}_k\}.
    \]

    \item \textbf{Multiple imputation of covariates:}
    Fit $M$ imputation models on the training covariates and generate completed training covariate sets
    \[
      \tilde{\mathcal{X}}^{(-k,m)} = \mathrm{Imp}^{(-k,m)}\!\big(\mathcal{X}^{(-k)}\big),
      \qquad m=1,\dots,M.
    \]
    Apply the same $M$ trained imputation models to the validation covariates:
    \[
      \tilde{\mathcal{X}}^{(k,m)} = \mathrm{Imp}^{(-k,m)}\!\big(\mathcal{X}^{(k)}\big),
      \qquad m=1,\dots,M.
    \]
    Define completed datasets by attaching outcomes:
    \[
      \tilde{\mathcal{D}}^{(-k,m)}
      = \{(\tilde{\boldsymbol{x}}_i^{(-k,m)}, y_i): i \notin \mathcal{I}_k\},\qquad
      \tilde{\mathcal{D}}^{(k,m)} 
      = \{(\tilde{\boldsymbol{x}}_i^{(k,m)}, y_i): i \in \mathcal{I}_k\}.
    \]
    \emph{(Tilde $\tilde\cdot$ indicates objects containing imputed values.)}

    \item \textbf{Centering:}  

    Compute the grand training mean for each covariate $j=1,\dots,p$ by averaging over all
    imputed training values across imputed datasets:
    \[
    \mu_j^{(-k)} 
    := \frac{1}{M|\mathcal{D}^{(-k)}|}
    \sum_{m=1}^M \sum_{i \notin \mathcal{I}_k} \tilde{x}_{ij}^{(-k,m)}.
    \]
    Center the covariates of the training and validation sets component-wise using these grand training means:
    \[
    \overline{x}_{ij}^{(-k,m)} = \tilde{x}_{ij}^{(-k,m)} - \mu_j^{(-k)}, 
    \qquad 
    \overline{x}_{ij}^{(k,m)}  = \tilde{x}_{ij}^{(k,m)}  - \mu_j^{(-k)},
    \]
    for all \(i\), all \(m\), and all \(j = 1, \dots, p\).

    Define the centered--imputed datasets
    \[
      \overline{\mathcal{D}}^{(-k,m)} := \{(\overline{\boldsymbol{x}}_i^{(-k,m)}, y_i): i \notin \mathcal{I}_k\},
      \qquad
      \overline{\mathcal{D}}^{(k,m)} := \{(\overline{\boldsymbol{x}}_i^{(k,m)}, y_i): i \in \mathcal{I}_k\}.
    \]
    \emph{(The bar $\overline{\cdot}$ denotes centered covariates or datasets composed of them. For clarity, the tilde indicating imputation is omitted in these barred symbols; they represent centered versions of the previously imputed values.)}

    \item \textbf{Run MIBoost and validate (for $t = 1,\dots,t_{\mathrm{stop}}$):}
    \begin{enumerate}[label*=\arabic*., leftmargin=*, labelsep=0.5em, itemsep=0pt, topsep=0pt]

      \item \textbf{Fit boosting paths on the centered training fold:}  
      Run \(\mathrm{MIBoost}\) for \(t_{\mathrm{stop}}\) iterations on the collection of
      centered--imputed training datasets \(\{\overline{\mathcal{D}}^{(-k,m)}\}_{m=1}^M\). For each
      \(m=1,\dots,M\), retain the imputation-specific predictor path
      \[
        \left\{\hat{\eta}^{(-k,m)[t]}(\overline{\boldsymbol{x}})\right\}_{t=1}^{t_{\mathrm{stop}}}.
      \]

      \item \textbf{Average additive predictors across imputations (per fold):}
      \[
        \overline{\eta}^{(-k)[t]}(\overline{\boldsymbol{x}})
        := \frac{1}{M}\sum_{m=1}^M \hat{\eta}^{(-k,m)[t]}(\overline{\boldsymbol{x}}).
      \]

      \item \textbf{Prediction and error computation on validation:}  
      For each imputed validation subset, predict with \(\overline{\eta}^{(-k)[t]}\):
      \[
        \hat{y}_i^{(k,m)}[t]
        = \overline{\eta}^{(-k)[t]}\!\big(\overline{\boldsymbol{x}}_i^{(k,m)}\big),
        \quad i \in \mathcal{I}_k,
      \]
      compute the validation error (per chosen loss) for each \(m\), and average these errors across \(m\)
      to obtain the fold-$k$ error at iteration \(t\).

    \end{enumerate}
\end{enumerate}

\noindent\rule[0.5ex]{\linewidth}{0.4pt}
\textbf{Post-processing across folds:}  
\begin{enumerate}[leftmargin=*]
    \setcounter{enumi}{4}
    \item \textbf{Average prediction errors across folds:}  
    For each iteration \(t\), average the validation errors across all \(K\) folds to obtain \(\mathrm{CV\;error}(t)\).

    \item \textbf{Select optimal stopping iteration:}  
    Choose 
    \[
      t_{\mathrm{stop}}^* = \arg\min_{t \leq t_{\mathrm{stop}}} \mathrm{CV\;error}(t).
    \]

    \item \textbf{Final model fitting:}  
    Impute the full covariate set \(\mathcal{X}\) \(M\) times to obtain
    \[
      \tilde{\mathcal{X}}^{(m)}=\mathrm{Imp}^{(m)}(\mathcal{X}),\qquad m=1,\dots,M,
    \]
    and define the completed datasets by attaching outcomes,
    \[
      \tilde{\mathcal{D}}^{(m)}=\{(\tilde{\boldsymbol{x}}_i^{(m)}, y_i)\}_{i=1}^n.
    \]
Compute the grand mean for each covariate \(j=1,\dots,p\) by averaging the within-imputation means,
\[
  \mu_j^{\mathrm{grand}}
  := \frac{1}{M}\sum_{m=1}^M \left(\frac{1}{n}\sum_{i=1}^n \tilde{x}_{ij}^{(m)}\right).
\]
Center all imputed datasets component-wise using these grand means, i.e.
\[
  \overline{x}_{ij}^{(m)} = \tilde{x}_{ij}^{(m)} - \mu_j^{\mathrm{grand}},
  \qquad i=1,\dots,n,\quad m=1,\dots,M,\quad j=1,\dots,p,
\]
and denote the resulting centered--imputed datasets by \(\overline{\mathcal{D}}^{(m)}\) (cf.\ Step~3). Fit \texttt{MIBoost} for
    \(t_{\mathrm{stop}}^*\) iterations on the collection \(\{\overline{\mathcal{D}}^{(m)}\}_{m=1}^M\), and retain the
    imputation-specific predictors \(\hat{\eta}^{(m)[t_{\mathrm{stop}}^*]}(\overline{\boldsymbol{x}})\), \(m=1,\dots,M\).
Define the final predictor as
\[
  \overline{\eta}^{[t_{\mathrm{stop}}^*]}(\overline{\boldsymbol{x}})
  := \frac{1}{M}\sum_{m=1}^M \hat{\eta}^{(m)[t_{\mathrm{stop}}^*]}(\overline{\boldsymbol{x}})
  =: \hat{\eta}^{[t_{\mathrm{stop}}^*]}(\overline{\boldsymbol{x}}).
\]
\end{enumerate}
\end{tcolorbox}

\vspace{25pt}

\subsection{Simulation setting}
\label{subsec:sim_setting}

\paragraph{Setup}  
We compared the performance of our proposed algorithm with established methods for variable selection in multiply imputed datasets. First, we applied traditional component-wise gradient boosting independently to each imputed dataset and derived a single final model using an estimate-averaging approach. We also applied the \texttt{SaLASSO} and \texttt{SaENET} algorithms implemented in the \texttt{R} package \texttt{miselect}. Exact specifications for the latter two are provided in Appendix~\ref{subsec:specifications}.

\vspace{8pt}

We conducted a simulation study under two settings, varying the number of covariates \(p\) (50 vs.\ 100). For each setting, we generated 50 datasets with \(n = 500\) observations. In each round, the data with missing values were split into a training set (80\%) and a test set (20\%). Within each training set, all methods applied an additional round of 5-fold cross-validation, repeatedly partitioning the data into inner training and validation subsets to determine the optimal hyperparameter values. All computations were performed in \texttt{R}. The following hyperparameters were tuned:

\vspace{3pt}

\begin{itemize}[itemsep=0pt, topsep=0pt,leftmargin=2em]
    \item number of boosting iterations \(t_{\mathrm{stop}}\),  
    \item shrinkage parameter \(\lambda\) in SaLASSO and SaENET,  
    \item elastic net parameter \(\alpha\) in SaENET.  
\end{itemize}

\paragraph{Data-generating mechanism}
The outcome was generated from the linear model
\[
\mathbf{y} = \beta_0 \mathbf{1}_n + \mathbf{X}\boldsymbol{\beta} + \boldsymbol{\varepsilon},
\qquad
\beta_0 = 5,
\qquad
\boldsymbol{\varepsilon} \sim \mathcal{N}_n\!\bigl(\mathbf{0}, 5^2 \mathbf{I}_n\bigr).
\]

The design matrix \(\mathbf{X} \in \mathbb{R}^{n \times p}\) contained \(p\) covariates, with \(q=5\) informative variables and \(p \in \{50,100\}\). Writing
\[
\mathbf{X} = [\mathbf{X}_{\mathrm{info}}, \mathbf{X}_{\mathrm{noise}}],
\]
the first \(q=5\) coefficients were independently redrawn from \(\mathrm{Unif}(1,2)\) for each simulation setting and replication, while the remaining coefficients were fixed at zero, that is,
\[
\beta_j =
\begin{cases}
\text{independently redrawn from } \mathrm{Unif}(1,2), & j \leq q, \\[4pt]
0, & j > q.
\end{cases}
\]

For each observation \(i=1,\dots,n\),
\[
\mathbf{x}_{i,\mathrm{info}} \overset{\mathrm{i.i.d.}}{\sim} \mathcal{N}_q(\mathbf{0}, \Sigma),
\qquad
\Sigma_{jk} =
\begin{cases}
1, & j = k, \\[4pt]
\rho, & j \ne k,
\end{cases}
\qquad
\rho = 0.25,
\]
and
\[
\mathbf{x}_{i,\mathrm{noise}} \overset{\mathrm{i.i.d.}}{\sim} \mathcal{N}_{p-q}\!\bigl(\mathbf{0}, \mathbf{I}_{p-q}\bigr),
\]
independently of \(\mathbf{x}_{i,\mathrm{info}}\). Thus, the first five covariates were correlated and informative, whereas the remaining \(p-q\) covariates were independent standard normal noise variables.

\paragraph{Missingness mechanism and imputation}  
We introduced missing values in \(\mathbf{X}\) under a missing‑at‑random (MAR) mechanism. In addition to $\mathbf{y}$, the first two covariates, $\mathbf{X}_1$ and $\mathbf{X}_2$, were fully observed, with the latter two driving the missingness in all other variables ($\mathbf{X}_j,\; j > 2$):

\[
\mathbb{P}(R_{ij}=0 \mid X_{i1}, X_{i2}) = \mathrm{logit}^{-1}\big(\gamma_{0} + \gamma_{1} X_{i1} + \gamma_{2} X_{i2}\big),
\]

\vspace{3pt}

where \(R_{ij}=0\) indicates a missing value, \((\gamma_{1}, \gamma_{2})^{\mathsf{T}} = (0.75, -0.5)^{\mathsf{T}}\), and the intercept \(\gamma_{0}\) was chosen such that the missingness proportion was approximately 25\%. Missing values were imputed using multivariate imputation by chained
equations (MICE; R package \texttt{mice}; \cite{vanBuuren.2011})
with predictive mean matching \cite{Little.1988}, generating $m=10$ imputed datasets. For each
variable, only covariates with absolute Spearman correlation of $|\rho|\ge 0.10$
with the target variable were included in that variable’s imputation model.

\paragraph{Evaluation metrics}  

Performance metrics considered were:  

\begin{multicols}{2}  
\begin{itemize}[itemsep=0pt, topsep=2pt]  
    \item Mean squared prediction error on test sets  
    \item Average tuned hyperparameters  
    \item Mean number of selected variables  
    \item True positive proportion  
    \item True negative proportion  
\end{itemize}  
\end{multicols}  

The true positive proportion is the proportion of informative covariates included in the final model, while the true negative proportion is the proportion of non-informative covariates excluded, both averaged across all simulation rounds. For both measures, values closer to 1 indicate better performance. 

\section{Results and Discussion}
\label{sec:results_discussion}

\subsection{Simulation results}
\label{sec:sim_results}

\begin{table}[!ht]
\centering
\begin{threeparttable}
\caption{\textbf{Simulation results.}}
\label{tab:results}
\begin{tabular}{c|l *{6}{c}}
\toprule
$p$ & Method & \textbf{MSPE} & $\lambda^*/\alpha^*$ & $t_{\mathrm{stop}}^*$ & \textbf{TPP} & \textbf{TNP} & \textbf{\# Selec.} \\
\midrule
\multirow{4}{*}{50} & \texttt{EA-Boosting} & 28.60 &  & 88.2 & 1.00 & 0.37 & 33.2 \\
                     & \texttt{MIBoost} & 28.60 &  & 81.8 & 1.00 & 0.80 & 14.0 \\
                     & \texttt{SaLASSO} & 29.16 & 4.0e-04 &  & 1.00 & 0.79 & 14.4 \\
                     & \texttt{SaENET} & 29.18 & 4.3e-04/7.9e-01 &  & 1.00 & 0.76 & 15.6 \\
\cmidrule(lr){1-8}
\multirow{4}{*}{100} & \texttt{EA-Boosting} & 29.83 &  & 79.9 & 1.00 & 0.59 & 44.2 \\
                     & \texttt{MIBoost} & 29.84 &  & 71.8 & 1.00 & 0.90 & 14.3 \\
                     & \texttt{SaLASSO} & 30.29 & 6.5e-04 &  & 0.99 & 0.89 & 15.3 \\
                     & \texttt{SaENET} & 30.46 & 6.1e-04/8.3e-01 &  & 0.99 & 0.87 & 17.4 \\
\bottomrule
\end{tabular}
\begin{tablenotes}
\small
\item \textit{Legend:} MSPE = Mean Squared Prediction Error; TPP = True Positive Proportion; TNP = True Negative Proportion; \# Selec. = Number of selected variables; EA = Estimate Averaging.
\end{tablenotes}
\end{threeparttable}
\end{table}

Prediction accuracy was comparable across all evaluated methods (Table \ref{tab:results}). There were almost no differences between gradient boosting involving estimate averaging and \texttt{MIBoost}.

In terms of variable-selection performance, all algorithms correctly identified the informative variables but also tended to include uninformative covariates. The model complexities and true negative proportions of \texttt{MIBoost} were similar to those of \texttt{SaLASSO} and \texttt{SaENET}. In contrast, gradient boosting with estimate averaging consistently tended to produce considerably more complex models and included considerably more uninformative covariates.

\subsection{Discussion}
\label{subsec:discussion}

In our simulation study, \texttt{MIBoost} showed prediction performance similar to that of \texttt{SaLASSO} and \texttt{SaENET} introduced in \cite{Du.2022}, as well as boosting with estimate averaging, but resulted in less complex models than the latter. It is unsurprising that predictive performance remained similar despite these differences in complexity, as accuracy is often only marginally affected by parsimony once the key predictors are included \cite{Wood.2008}. However, it should be noted that estimate averaging led to the inclusion of substantially more covariates. In practice, the application of such models would require considerably more information, potentially resulting in a more laborious and costly data collection process. Moreover, when model interpretability is a priority, increased complexity inevitably reduces transparency and complicates interpretation.

Furthermore, it is important to emphasize that we introduced \texttt{MIBoost} specifically in the context of component‑wise gradient boosting, which implicitly leads to variable selection,
since the need for a single interpretable model derived from multiply imputed datasets is particularly pronounced in statistical learning settings in which interpretability is essential. Therefore, we chose to compare \texttt{MIBoost} with \texttt{SaLASSO} and \texttt{SaENET}. Nonetheless, the underlying principle— selecting, across all imputed datasets, the one base-learner that minimizes an aggregated loss function — can be readily generalized to other boosting frameworks. Accordingly, the approach can be extended in the future to likelihood-based boosting, in which an aggregated likelihood can be defined, as well as to non-statistical boosting methods.

\section{Conclusion}
\label{sec:conclusion}

In this paper, we proposed \texttt{MIBoost}, a novel approach for applying component-wise gradient boosting to multiply imputed datasets. The method operates by minimizing an aggregated loss function across imputations, ensuring a uniform variable-selection mechanism such that, in every iteration, the same covariate is selected across all datasets. In a simulation study, \texttt{MIBoost} achieved predictive performance comparable to that of other established methods designed to ensure uniform variable selection after multiple imputation. Importantly, the underlying principle of the algorithm can be readily generalized to other boosting frameworks in future work. To facilitate its application in practice, both \texttt{MIBoost} and its cross-validation framework \texttt{MIBoostCV} are available in the open-source \texttt{R} package \texttt{booami}.

\section*{Declarations}

\noindent\textbf{Funding} \\
This research did not receive any specific grant from funding agencies in the public, commercial, or not-for-profit sectors.

\vspace{0.5em}
\noindent\textbf{Declaration of Competing Interest} \\
The authors declare that they have no known competing financial interests or personal relationships that could have appeared to influence the work reported in this paper.

\vspace{0.5em}
\noindent\textbf{Data availability}
No real data were generated or analyzed during the current study; all analyses are based on simulated data. The R code used for the simulation study is publicly available at the GitHub repository associated with this article, \url{https://github.com/RobertKuchen/miboost-code}, and archived on Zenodo at \url{https://doi.org/10.5281/zenodo.19322771}.

\vspace{0.75em}
\noindent\textbf{Author contributions}
\medskip

\begin{tabular}{@{}p{3.8cm}p{\dimexpr\linewidth-3.8cm\relax}@{}}
\textbf{Robert Kuchen:} & Conceptualization, Methodology, Software, Formal analysis, Writing -- original draft.\\[0.35em]
\textbf{Noemi Castelletti:} & Supervision, Writing -- review \& editing.\\[0.35em]
\textbf{Konstantin Strauch:} & Supervision, Writing -- review \& editing.
\end{tabular}

\medskip
\noindent All authors read and approved the final manuscript.

\bibliography{MIBoost_references}

\appendix
\section*{Appendix}
\label{sec:appendix}
\addcontentsline{toc}{section}{Appendix}

% --- Keep "Appendix" unnumbered, but label appendix subsections as A, B, C
\setcounter{subsection}{0}
\renewcommand\thesubsection{\Alph{subsection}}

\setcounter{subsubsection}{0}
\renewcommand\thesubsubsection{\thesubsection.\arabic{subsubsection}}

\subsection{Existing approaches for variable selection with multiply imputed data}
\label{subsec:existing_approaches}

Several strategies exist for fitting prediction models on multiply imputed datasets. The simplest is \textit{naive data stacking} \cite{Beesley.2021}, in which all \(M\) imputed datasets are combined into one large dataset comprising \(M \times n\) observations. While easy to implement, this approach artificially inflates the sample size, possibly resulting in overfitting. 

An alternative approach is \textit{estimate averaging} \cite{Wood.2008}, wherein model selection is conducted independently within each imputed dataset, and the $M$ resulting sets of coefficient estimates are subsequently averaged. However, as variables selected in one dataset may be excluded in another, the resulting model may contain a substantial number of non-zero coefficients, thereby compromising model parsimony.

To address this, \textit{selection frequency thresholding} \cite{Zhao.2017} has been proposed. In this approach, only variables selected in at least a prespecified proportion of imputations are retained, with a 50\% cutoff (“majority vote”) being most common. Other thresholds (e.g., 25\% or 40\%) can be used to balance sparsity and stability.  While this approach leads to sparser models than estimate averaging, the cutoff is arbitrary and relevant variables with unstable signals may be excluded. Moreover, dependencies between covariates are partially ignored, and imputation uncertainty is not fully captured. 

Beyond these, more advanced approaches have been proposed. 
\textit{Bayesian data-augmentation} methods \cite{Tanner.1987} 
integrate over the missing-data distribution within a unified 
posterior framework, naturally propagating uncertainty. 
However, they often entail high computational demands and 
can be sensitive to the choice of prior distributions. \textit{Bootstrapped inclusion-probability methods} \cite{Sauerbrei.1992} use resampling across imputations to stabilize variable selection but require extensive computation and lack clear thresholds for inclusion. \textit{EM-based penalized regression} \cite{Chen.2014} integrates imputation with regularized model fitting, offering efficiency but relying on strong distributional assumptions and custom model-selection criteria. Finally, \textit{joint modeling approaches} \cite{JosephL.Schafer.1997} fit a single model across all imputations by combining likelihoods, although these methods can be complex to implement and are less explored in practice.  

In recent years, two novel post-imputation variable-selection frameworks---\allowbreak
grouping and stacked methods---\allowbreak
have been proposed to promote uniform variable selection across multiply imputed datasets. In the grouping framework, the group LASSO \cite{Yuan.2006} has been applied to post-imputation variable selection \cite{Chen.2013}. Although all imputed datasets remain independent, leading to different coefficient estimates in each, coefficients for the same covariate are penalized jointly, resulting in uniform variable selection across imputed datasets. In the stacked framework \cite{Wood.2008}, the overfitting risk associated with combining imputed datasets into a single augmented dataset is addressed by down-weighting observations. It was suggested to fit the elastic net \cite{Zou.2005} to a single stacked dataset, with each subject’s observations weighted so that their total weight sums to one, yielding a single final model without inflating the sample size \cite{Wan.2015}.

These approaches were extended and provided with a user-friendly software implementation in \cite{Du.2022}. In both frameworks, the \textit{standard} and \textit{adaptive} versions of LASSO \cite{Zou.2006} and the elastic net \cite{Zou.2009} can be applied, resulting in a total of six models. Since the stacked adaptive formulations of LASSO and the elastic net (denoted \texttt{SaLASSO} and \texttt{SaENET}) achieved the best performance in \cite{Du.2022}, we focus on these two in the comparison with our approach, \texttt{MIBoost}. \texttt{SaLASSO} and \texttt{SaENET}, implemented in the \texttt{R} package \texttt{miselect} \cite{Rix.2020}, rely on minimizing a single loss function, as does \texttt{MIBoost}.

\subsection{Statistical and Component-Wise Gradient Boosting}
\label{subsec:background_gradient_boosting}

Statistical gradient boosting \cite{JeromeH.Friedman.2001} is a machine learning algorithm that minimizes a specified differentiable loss function $\rho$ by incrementally constructing an additive model in a stagewise manner \cite{TrevorHastie.1986,JeromeFriedman.2000}. 
Given $n$ observations, we define the \textit{empirical loss function} as
\[
\hat{R}(\hat{\bm{\eta}}(\boldsymbol{X})) 
= \sum_{i=1}^n \rho\big(y_i, \hat{\eta}(\boldsymbol{x}_i)\big),
\]
which quantifies the discrepancy between the additive predictor $\hat{\eta}(\boldsymbol{X})$ and the observed responses $\boldsymbol{y}$. 
\[
\hat{\boldsymbol{\eta}}^{[t]}(\boldsymbol{X})
:= \big(\hat{\eta}^{[t]}(\boldsymbol{x}_1),\dots,\hat{\eta}^{[t]}(\boldsymbol{x}_n)\big)^{\top}\in\mathbb{R}^n.
\]
When evaluated at the fitted predictor $\hat{\eta}(\boldsymbol{X})$, this yields the empirical loss $\hat{R}(\hat{\eta}(\boldsymbol{X}))$. 
Boosting achieves minimization of this loss by iteratively descending $\hat{R}(\hat{\eta}(\boldsymbol{X}))$ along the direction of its negative gradient. Specifically, at each iteration, we compute
\[
\bm{u}^{[t]}
:= - \left.\nabla_{\bm{\eta}} \hat{R}(\bm{\eta})\right|_{\bm{\eta}=\hat{\bm{\eta}}^{[t-1]}(\boldsymbol{X})},
\qquad \bm{u}^{[t]} \in \mathbb{R}^n,
\]
which corresponds to the negative gradient of the empirical loss with respect to the additive predictor, evaluated at its current value (often referred to as \textit{pseudo-residuals}). 
In the special case of squared error loss,
\[
\rho(y, \eta) = \tfrac{1}{2}(y - \eta)^2,
\]
these pseudo-residuals coincide with the ordinary residuals,
\[
\boldsymbol{u}^{[t]} = \boldsymbol{y} - \hat{\boldsymbol{\eta}}^{[t-1]}(\boldsymbol{X}),
\qquad \boldsymbol{u}^{[t]} \in \mathbb{R}^n.
\]

\subsubsection*{Design matrix and base-learners}

The additive predictor $\hat{\eta}(\boldsymbol{X})$ is a function of the design matrix $\boldsymbol{X}$, which is constructed from a set of $c$ raw covariates (e.g., \texttt{age}, \texttt{sex}). 
From these covariates, we derive a collection of $p$ \textit{candidate covariates} ($c \leq p$), which form the columns of the design matrix,
\[
\boldsymbol{X} = [\boldsymbol{X}_1, \dots, \boldsymbol{X}_p] = 
\begin{bmatrix}
\boldsymbol{x}_1^T \\
\vdots \\
\boldsymbol{x}_n^T
\end{bmatrix},
\]
where $\boldsymbol{X}_j$ denotes the vector of values of the $j$-th candidate covariate across all observations, and $\boldsymbol{x}_i$ the $i$-th row vector containing all candidate covariate values for observation $i$. 
For convenience, we use the shorthand $\eta_i := \eta(\boldsymbol{x}_i)$ for the additive predictor of observation $i$. 
Candidate covariates may be simple transformations (e.g., \texttt{age}, \texttt{age}$^2$) or more flexible effects, such as penalized splines or random intercepts. 
To ensure identifiability and comparability, candidate covariates (i.e., their design matrices) are typically centered so that their contribution is orthogonal to the intercept. 
Building on these candidate covariates, we define \textit{base-learners} as simple prediction rules (e.g., univariate regressions or smoothers) that estimate the association of each candidate covariate with the outcome.

In this paper we focus on \textit{component-wise gradient boosting} \cite{Buhlmann.2003}, in which each base-learner corresponds only to a single candidate covariate. 
Furthermore, at each iteration, only a single base-learner is selected and updated, while all others remain unchanged. 
Formally, at each boosting iteration, denoted by $t$ in the following, we compute, for each of the $p$ base-learners, the \textit{least-squares projection} of the current negative gradient vector $\bm{u}^{[t]}$ (the pseudo-residuals) onto the function space of that base-learner:
\[
\hat{h}_{\beta_r}^{[t]} = \arg\min_{h \in \mathcal{H}_r} \sum_{i=1}^n \big( u_i^{[t]} - h(x_{i,r}) \big)^2,
\qquad \hat{h}_{\beta_r}^{[t]}: \mathbb{R} \to \mathbb{R},
\]
where $\mathcal{H}_r$ denotes the function space associated with base-learner $r$. 
The estimated function $\hat{h}_{\beta_r}^{[t]}$ is referred to as the \textit{empirical base-learner} for candidate covariate $r$ in iteration $t$. 

For a simple linear base-learner, this corresponds to an ordinary least-squares regression of the pseudo-residuals on an intercept and the $r$\hyp{}th covariate. 
This yields the fitted function
\[
\hat{h}_{\beta_r}^{[t]}(x_r) 
   = \hat{\beta}_{0,r}^{[t]} + \hat{\beta}_r^{[t]} x_r,
   \qquad \hat{h}_{\beta_r}^{[t]}: \mathbb{R} \to \mathbb{R}.
\]

Evaluating this function at all training observations gives the vector of fitted values
\[
\hat{\boldsymbol{h}}_{\beta_r}^{[t]}(\boldsymbol{X}_r)
   = \hat{\beta}_{0,r}^{[t]} \boldsymbol{1} 
   + \hat{\beta}_r^{[t]} \boldsymbol{X}_r,
   \qquad \hat{\boldsymbol{h}}_{\beta_r}^{[t]}(\boldsymbol{X}_r) \in \mathbb{R}^n,
\]
where $\boldsymbol{1}$ is the $n$-dimensional vector of ones and $\boldsymbol{X}_r$ is the $n$-dimensional vector containing the $r$\hyp{}th covariate across all observations.

We define the residual sum of squares for base-learner $r$ in iteration $t$ as
\[
L_r^{[t]} := \sum_{i=1}^n \big(u_i^{[t]} - \hat{h}_{\beta_r}^{[t]}(x_{i,r})\big)^2,
\]
and select the base-learner with minimal loss,
\[
r^* = \arg\min_r L_r^{[t]}.
\]

The additive predictor is then updated by moving a small step $\nu$ (typically $\nu = 0.1$) in the direction of the selected base-learner. 
In function notation, this reads
\[
\hat{\eta}^{[t]}(\boldsymbol{x}) 
= \hat{\eta}^{[t-1]}(\boldsymbol{x}) 
+ \nu \,\hat{h}_{\beta_{r^*}}^{[t]}(x_{r^*}),
\qquad \hat{\eta}^{[t]} : \mathbb{R}^p \to \mathbb{R}.
\]

Equivalently, when evaluated at all training points, the update can be written in vector form as
\[
\hat{\boldsymbol{\eta}}^{[t]}(\boldsymbol{X})
= \hat{\boldsymbol{\eta}}^{[t-1]}(\boldsymbol{X}) 
+ \nu \hat{\boldsymbol{h}}_{\beta_{r^*}}^{[t]}(\boldsymbol{X}_{r^*}),
\qquad \hat{\boldsymbol{\eta}}^{[t]}(\boldsymbol{X}) \in \mathbb{R}^n.
\]

Note that the same base-learner can be selected multiple times. Consequently, with only one base-learner updated per iteration, a small number of iterations before stopping $t_{\mathrm{stop}}$ typically leaves some covariates excluded and, due to the small step length, results in coefficient shrinkage. This makes component-wise gradient boosting conceptually related to penalized regression approaches such as the LASSO and elastic nets \cite{Tibshirani.1996,Buhlmann.2003,PeterBuhlmann.2006}. 

Moreover, unlike black-box machine-learning methods, the resulting additive model remains interpretable and can flexibly accommodate different outcome types—including continuous, binary, count, and time-to-event data—by specifying an appropriate loss function $\rho$. These properties make component-wise boosting particularly attractive for high-dimensional problems that require balancing predictive performance with model sparsity.

The optimal number of iterations, denoted by $t_{\mathrm{stop}}^*$, is determined by identifying the number of iterations that results in the model with the lowest prediction error. Typically, this number is selected using cross-validation (CV).

\subsection{Specifications for \texttt{SaLASSO} and \texttt{SaENET}}
\label{subsec:specifications}

First of all, it is important to note that the \texttt{miselect} package provides built-in cross-validation (CV) functions to find the optimal values for \(\lambda\) and \(\alpha\). However, these functions can only be applied to data that have already been imputed. As elucidated in subsection \ref{subsec:MIBoostCV}, to prevent data leakage, data should only be imputed \emph{after} being split into the respective training and validation subsets. We therefore wrote our own code to split the data, perform imputation, and carry out 5-fold CV.

To find the optimal $\lambda$ and $\alpha$ values, we first needed to define a $\lambda$ sequence for cross-validation. To obtain such a sequence, we initially ran both \texttt{SaLASSO} and \texttt{SaENET} for 50 preliminary simulation rounds using an automatically generated sequence of 100 values for $\lambda$, spaced logarithmically from $\lambda_{\max}$ (the smallest value for which all coefficients are zero) down to $\lambda_{\min} = 10^{-6} \lambda_{\max}$. We then recorded the optimal $\lambda$ values obtained across the CV folds. Based on these results, we constructed a new $\lambda$ sequence of length 100 that extended slightly beyond the observed range of optimal values by setting its lower bound to half the smallest selected value and its upper bound to twice the largest selected value. The sequence was again generated on a logarithmic scale, thereby allocating a higher density of values in the region in which most optimal $\lambda$ values were concentrated.

For \texttt{SaENET} only, cross-validation was also used to select an optimal value for \(\alpha\). To that end, we created an equally spaced \(\alpha\) sequence from 0 to 1, consisting of 41 values.  

Having determined the \(\lambda\) and \(\alpha\) sequences for cross-validation, we ran our custom CV code, which implemented the functions \texttt{galasso} and \texttt{saenet}, but wrapped them in a framework that first split the data into training and validation sets and then performed imputation. For \texttt{SaLASSO}, in each simulation round we selected the \(\lambda\) value associated with the smallest CV error. For \texttt{SaENET}, we performed a grid search over all \(41 \times 100 = 4100\) \((\lambda, \alpha)\) pairs and selected the pair with the smallest CV error.

Furthermore, the \texttt{miselect} package allows for weighting observations to account for the fact that stacked datasets consist of \(M \times n\) observations. For both \texttt{SaLASSO} and \texttt{SaENET}, we chose observation weights of \(1/M\), as recommended in \cite{Du.2022}. An alternative weighting approach—scaling each observation by the proportion of missing values across all variables—was also discussed in \cite{Du.2022}. However, the authors acknowledged that this method would shift the analysis closer to a complete-case analysis, which runs counter to the purpose of multiple imputation. We therefore did not adopt this approach.

Finally, since we did not apply standard LASSO or elastic nets but rather their adaptive versions, we needed to define adaptive weight vectors for each run of \texttt{SaLASSO} or \texttt{SaENET}. To do so, we first fitted a preliminary linear regression model with equal weights. We then extracted the estimated coefficients for all candidate covariates and used them to compute the corresponding adaptive weight vectors.

\end{document}